\documentclass[journal]{IEEEtran}

\usepackage[utf8]{inputenc}
\usepackage[T1]{fontenc}
\usepackage{cite}
\usepackage{amsmath,amssymb}
\usepackage{graphicx}
\usepackage{booktabs}
\usepackage{makecell}
\usepackage{siunitx}
\DeclareSIUnit{\pixel}{px}
\DeclareSIUnit{\frame}{frame}

\title{Comprehensive Evaluation of Rule-Based, Machine Learning, and Deep Learning in Human Estimation Using Radio Wave Sensing: Accuracy, Spatial Generalization, and Output Granularity Trade-offs
}

\author{
    Tomoya Tanaka$^{1,2}$, 
    Tomonori Ikeda$^{2}$, 
    and Ryo Yonemoto$^{2}$%
    \thanks{$^{1}$Tomoya Tanaka is with the School of Electrical and Computer Engineering, 
    Georgia Institute of Technology, Atlanta, GA 30332 USA (e-mail: tomoya.tanaka@gatech.edu).}%
    \thanks{$^{2}$Tomoya Tanaka, Tomonori Ikeda, and Ryo Yonemoto are with SoftBank Corp., 
    Tokyo, Japan (e-mail: \{tomoya.tanaka, tomonori.ikeda, ryo.yonemoto\}@softbank.jp).}%
}

\begin{document}

\maketitle
\thispagestyle{empty}
\pagestyle{empty}

\begin{abstract}
This study presents the first comprehensive comparison of rule-based methods, traditional machine learning (ML) models, and deep learning (DL) models in radio-wave sensing with frequency-modulated continuous wave (FMCW) multiple-input multiple-output (MIMO) radar. We systematically evaluated five approaches in two indoor environments with distinct layouts: a rule-based connected-component method; three traditional ML models, namely k-nearest neighbors (KNN), random forest (RF), and support vector machine (SVM); and a DL model combining a convolutional neural network (CNN) and long short-term memory (LSTM). In the training environment, the CNN--LSTM achieved the highest accuracy, while traditional ML models provided moderate performance. In a new layout, however, all learning-based methods showed significant degradation (drops of 13.02--20.73 percentage points), whereas the rule-based method remained stable. Notably, for binary detection (presence vs. absence of people), all models consistently achieved high accuracy across layouts. These results demonstrate that high-capacity models can produce fine-grained outputs with high accuracy in the same environment, but they are vulnerable to domain shift. In contrast, rule-based methods cannot provide fine-grained outputs but exhibit robustness against domain shift. Moreover, regardless of the model type, a clear trade-off was revealed between spatial generalization performance and output granularity.
\end{abstract}



\section{INTRODUCTION}

Radio wave sensing systems utilizing Frequency Modulated Continuous Wave (FMCW) radar have attracted significant attention as a promising alternative to camera-based methods for human estimation tasks in indoor environments, owing to their privacy-preserving characteristics, robustness to lighting conditions, and penetration capabilities~\cite{Kong2025,millimeterReview}. In particular, when applying deep learning (DL) techniques, Multiple-Input Multiple-Output (MIMO) configurations are widely employed to enhance spatial resolution. These systems have been applied in elderly monitoring, occupancy detection, smart buildings, and robotics~\cite{frequencyTracking,SensorsVehOcc2024,SmartBldgMultiRadar2024,radarays}.

Human estimation methods have progressed from early rule-based approaches with high interpretability but limited accuracy~\cite{adaptiveThresholding}, to traditional machine learning (ML) models such as k-nearest neighbors (KNN), support vector machine (SVM), and random forest (RF), and further to deep learning (DL) models combining convolutional neural networks (CNNs) and long short-term memory (LSTM), as well as attention/transformer-based architectures, which have achieved notable accuracy improvements under controlled environments~\cite{humanMotion,hybridCNN,deepLearning}.

However, ML/DL models exhibit substantial performance degradation when deployed in environments with spatial configurations different from those in the training data. This is caused by domain shift, where input distributions change due to furniture rearrangement, sensor aging, or environmental variations, violating the i.i.d. assumption. Recent cross-environment studies quantitatively report large drops in recognition performance under such shifts~\cite{CrossEnvElectronics2025}. Our prior work has also demonstrated that multiple ML models suffer accuracy degradation under environmental changes~\cite{ourPriorWork}, whereas simpler rule-based and thresholding approaches can maintain relatively stable behavior in some settings~\cite{adaptiveThresholding}.

Existing studies on radio wave sensing have the following limitations.  
(1) \emph{Lack of evaluation metrics for spatial generalization} — Most prior works are limited to accuracy verification within the same environment, and have not systematically employed quantitative metrics to evaluate generalization performance under changes such as furniture rearrangement or application to new environments~\cite{microwaveSurveyCorrect,passiveRadar}.  
(2) \emph{Limited scope of domain shift evaluation} — Although a few studies, such as Hernangómez et al.~\cite{Hernangomez2022} and Khodabakhshandeh et al.~\cite{Khodabakhshandeh2021}, have investigated domain adaptation across FMCW radar configurations, these efforts are largely restricted to deep learning–based approaches focusing on single-model scenarios. No comprehensive comparisons including rule-based methods and conventional machine learning models have been conducted.  
(3) \emph{Unexplored relationship between output granularity and spatial generalization} — While presence/occupancy detection and multi-class people counting have both been studied~\cite{GroupedPeopleCounting2023,SmartBldgMultiRadar2024}, no prior work has systematically investigated how different output granularities (e.g., four-class people counting vs.\ binary presence detection) affect robustness against domain shift.

In summary, no study in radio wave sensing has provided a unified and comprehensive evaluation that simultaneously considers accuracy, spatial generalization, and output granularity. In this work, we evaluate five approaches (rule-based, KNN, SVM, RF, CNN--LSTM) under unified conditions across two environments with different furniture configurations (0--4 items vs.\ 6 items), systematically analyzing the trade-offs among accuracy, spatial generalization, and output granularity. This study constitutes the first comprehensive analysis of methods and performance in radio wave sensing, and the findings represent an important milestone toward accelerating the transition of these systems from research prototypes to practical deployment.

\section{EXPERIMENTAL SETUP}

The radar hardware, environment setup, and data acquisition used in this study are shared with our ongoing research on domain adaptation. In contrast, the present work focuses on evaluating algorithmic trade-offs across different layout configurations within the same physical environment.

\subsection{Radar System Overview}

Frequency-Modulated Continuous Wave (FMCW) radar is a representative sensing technique that estimates the distance and velocity of objects by analyzing the time-dependent frequency shift between transmitted and received signals~\cite{exactLinear}. By performing a linear sweep of the carrier frequency, FMCW radar systems generate beat signals, which can be transformed into range information through signal processing methods such as the Fast Fourier Transform (FFT).  

In this study, we utilized a millimeter-wave MIMO FMCW radar system operating at \SI{24.15}{\giga\hertz}. The system incorporates two transmitting antennas and four receiving antennas, and produces a range--azimuth amplitude map with a resolution of $12 \times 91$. This representation enables the simultaneous characterization of range and azimuth, where regions of markedly elevated amplitude values suggest the existence and spatial location of human subjects.  

The radar apparatus, experimental environment, and data collection procedures employed in this study are consistent with those used in our ongoing research on domain adaptation. For comprehensive information regarding hardware specifications, the signal processing architecture, and the design of measurement conditions, readers are referred to our prior work published in the IEICE Technical Report~\cite{ieiceRadarTechRep}.

\subsection{Environment Configuration}
Training and testing datasets were derived from two experimental conditions, labeled as Environment~A and Environment~B. Their detailed properties are presented in Table~I.

\begin{table}[htp]
\centering
\caption{Environment Specifications Comparison}
\label{tab:env_specs}
\renewcommand{\arraystretch}{1.2}
\begin{tabular}{p{2.2cm}|p{2.5cm}|p{2.5cm}}
\hline
\textbf{Specification} & \textbf{Environment A} & \textbf{Environment B} \\
\hline
Room Size & $\SI{4.9}{\meter} \times \SI{6.9}{\meter}$ & $\SI{4.9}{\meter} \times \SI{6.9}{\meter}$ \\
Ceiling Height & $\SI{2.7}{\meter}$ & $\SI{2.7}{\meter}$ \\
Wall Material & RF Absorber & RF Absorber \\
Floor Material & Carpet & Carpet \\
Major Furniture & 0–4 items & 6 items \\
Radar Position & (x, y, z) & (x, y, z) \\
Acoustic Property & Anechoic & Anechoic \\
\hline
\end{tabular}
\end{table}

\begin{figure}[htp]
\centering
\includegraphics[width=0.8\linewidth]{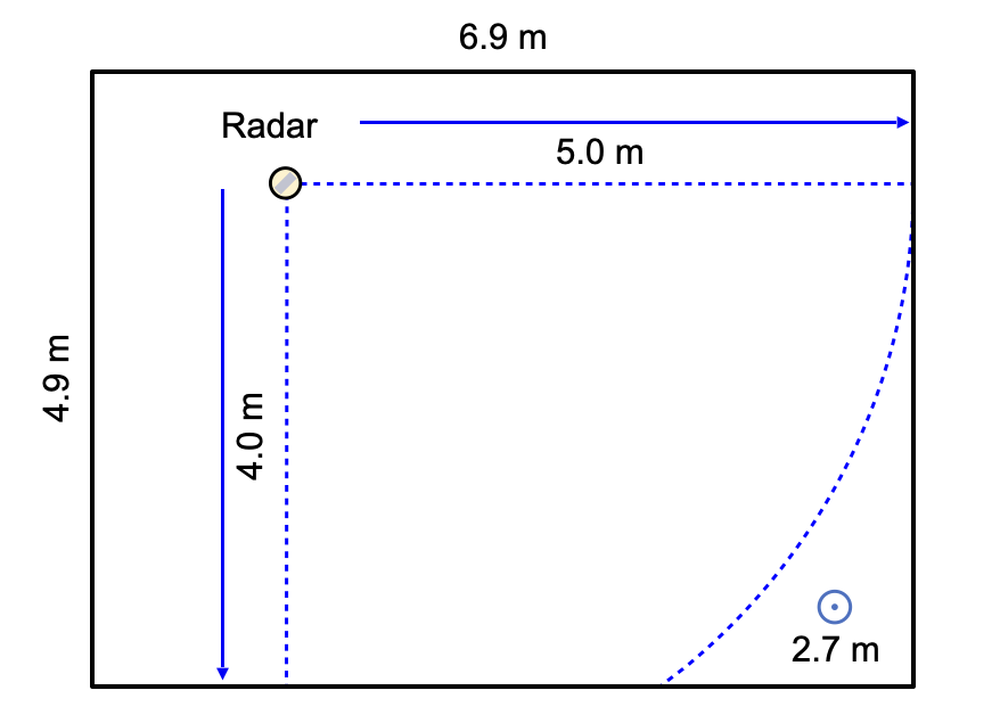}
\caption{Skeletal structure of the room in Environment A and B. The dashed lines indicate the coverage area within which objects can be detected by the radar.
}
\label{fig:envA}
\end{figure}

The unified skeletal structure of the room used for both Environments~A and~B is illustrated in Fig.~\ref{fig:envA}. This room is a radio anechoic chamber. For Environment~A, four simple configurations were prepared within the same room: (i) an empty setting with no furniture, (ii) a setting where between one and four chairs were placed at random, (iii) a setting where two desks were joined together and positioned at the center of the room, and (iv) a setting where a whiteboard was placed at the center. In contrast, Environment~B was designed with a more complex arrangement, consisting of three chairs, two desks, and one whiteboard. The purpose of defining these two environments within the same physical space was to examine how well model performance could be maintained under layout variations that frequently arise in practical deployment scenarios.  

In real-world usage, such rearrangements of furniture and equipment occur on a routine basis. Chairs are often moved to accommodate meetings, desks are repositioned for collaborative activities, and, over time, the overall room organization gradually changes in accordance with user preferences and functional requirements. Therefore, it is essential to evaluate whether algorithms can remain stable and effective under these inevitable and continuously occurring environmental changes.

\subsection{Dataset Overview}

Table~\ref{tab:dataset} reports the distribution of collected samples across Environments~A and~B.  

\begin{table}[htp]
\centering
\caption{Number of samples per class in each environment}
\label{tab:dataset}
\begin{tabular}{ccc}
\hline
\textbf{Class (Number of Persons)} & \textbf{Env A} & \textbf{Env B} \\
\hline
0 & 1600 & 400 \\
1 & 1600 & 400 \\
2 & 1600 & 400 \\
3 & 1600 & 400 \\
\hline
\end{tabular}
\end{table}

Each instance is represented by a three-dimensional tensor of size $12 \times 91 \times 60$. The first axis corresponds to range bins (radial distance from the radar), the second axis corresponds to angular bins (azimuthal direction), and the third axis represents the temporal sequence of frames. The entries of this tensor store amplitude values of the received signals.  

Data collection involved three male volunteers in their thirties and forties. Although the number of participants was modest, it was sufficient to address the central aim of this work, namely comparing the effectiveness of alternative algorithmic approaches. Because radar backscatter characteristics are largely determined by physical properties such as body dimensions and movement trajectories, the influence of inter-individual differences on algorithmic evaluation is expected to be minimal.  

Each sample belonged to one of the following activity categories, designed to mimic realistic usage contexts:  
\begin{enumerate}
    \item All subjects standing stationary.  
    \item All subjects walking in random trajectories within the radar coverage.  
    \item A combination of stationary and moving subjects.  
\end{enumerate}

For Environment~A, 1,600 samples were collected for each occupancy class (0–3 persons). Within each class, the data were evenly allocated across the four layout settings, yielding 400 samples per configuration. These samples were randomly divided into training, validation, and testing subsets with a split ratio of 7:1.5:1.5.  

For Environment~B, 400 samples were collected per class. These data were used solely for evaluating cross-environment generalization of models trained on Environment~A. 

\subsection{Data Preprocessing}

All collected radar data underwent an identical preprocessing routine prior to model training and assessment, so that measurement consistency could be preserved despite variations in environmental conditions. This procedure consisted of two main steps: outlier handling and normalization, as described below.

\textbf{Outlier Handling:} Extreme values in the amplitude data were removed by restricting the distribution to fall within the range defined by the 0.1st and 99.9th percentiles. Values outside this interval were clipped to the nearest boundary.

\textbf{Normalization:} After outlier handling, the data were scaled using min--max normalization. Specifically, the minimum was aligned to zero and the maximum to one, ensuring that all processed values were mapped onto a consistent scale between 0 and 1.

\textbf{Sigmoid-based Weighting}
In this study, we utilized two-dimensional amplitude maps with a resolution of $12 \times 91$ (range $\times$ azimuth). For each element of the map, the standard deviation along the temporal dimension was calculated, and based on this result, a weight map was generated using a sigmoid function to rescale the amplitude values. The weight is defined as

\begin{equation}
w(x, y) = \frac{1}{1 + \exp\!\left(-\frac{\sigma(x, y) - \tau}{s}\right)} .
\end{equation}

Traditionally, radar-based studies often employ squared amplitude (i.e., power) as the primary feature representation. In contrast, our approach still uses the amplitude values themselves as the input features, but rescales them through a weighting scheme derived from their temporal standard deviation. This weighting emphasizes amplitude components that exhibit human-related fluctuations, thereby highlighting subtle physiological variations and micro-movements while suppressing static background reflections. As a result, the procedure emphasized informative features while suppressing irrelevant fluctuations, and improvements in both accuracy and generalization were observed across all models. Consequently, the comparative behaviors of the models became clearer and more interpretable, and we incorporated this step as part of the preprocessing pipeline.

\section{Validation Models}

\subsection{Rule-Based Connected Component (CC) Model}

\subsubsection{Overview}
The Rule-Based Connected Component (CC) Model is a radar-based people counting approach that integrates temporal variation analysis (peak detection) widely used in signal processing with connected component analysis established in computer vision. This method overcomes the limitations of conventional fixed-threshold and single-parameter approaches by adopting an ensemble strategy with multiple thresholds and multiple windows, utilizing temporal change patterns in radar reflections for people counting.

\subsubsection{Theoretical Foundation}

\begin{figure}[htbp]
\centering
\includegraphics[width=1.0\columnwidth]{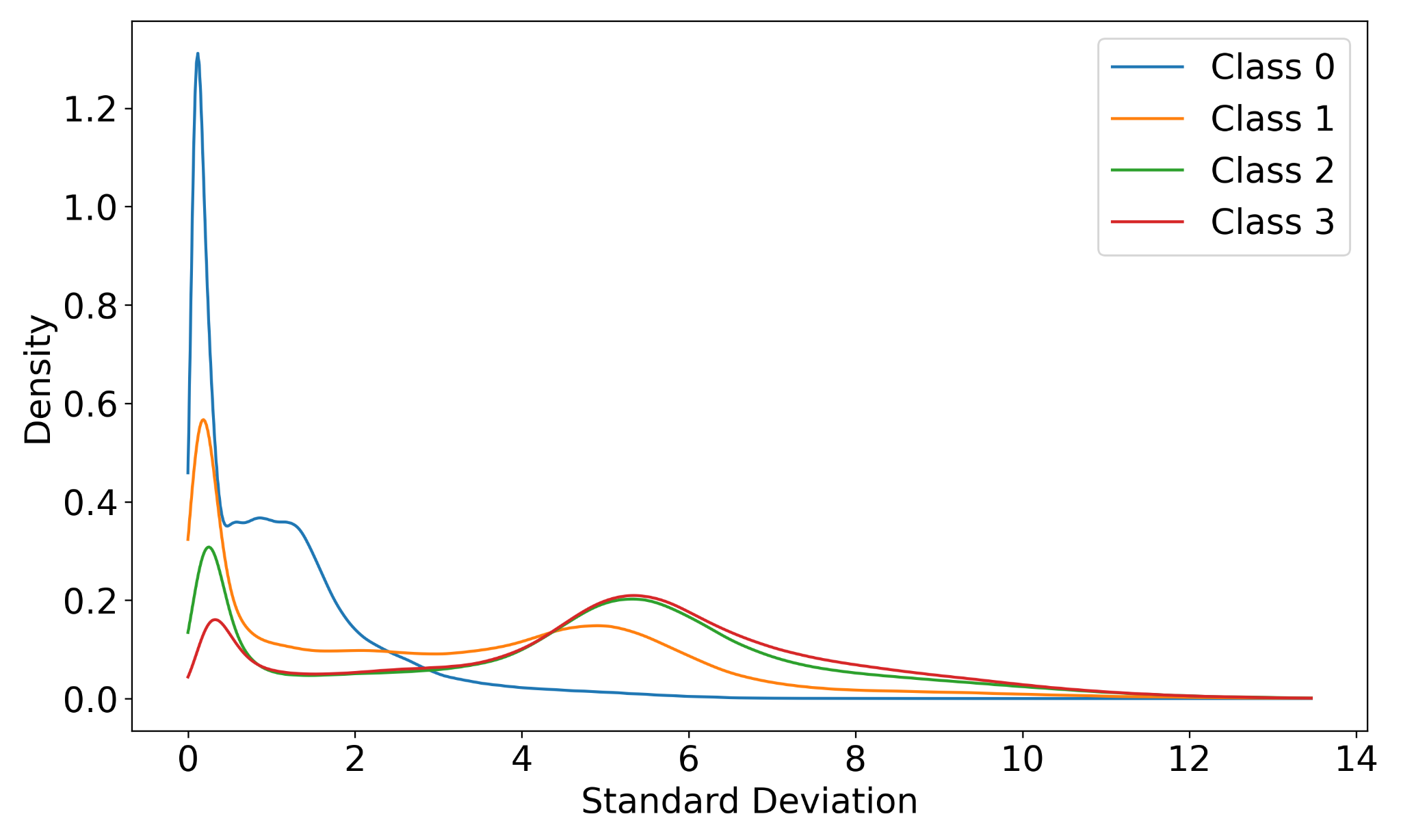}
\caption{Distribution of temporal standard deviation for different numbers of people. 
Blue line represents 0 persons, orange line represents 1 person, green line represents 2 persons, 
and red line represents 3 persons. This figure was generated from Environment~A, 
using 6,400 data samples (1,600 samples for each class).}
\label{fig:std_distribution}
\end{figure}

\paragraph{Differences in Standard Deviation Distribution by Number of People}
Figure~\ref{fig:std_distribution} shows histograms of temporal standard deviation by number of people. While 0 people (static environment) concentrates on low standard deviation values, 1-3 people show increasing frequency of high standard deviation values as the number of people increases. This difference in distribution characteristics forms the theoretical foundation of this method, and the overlapping portions of distributions provide the basis for multi-threshold settings.

\subsubsection{Overview of Processing Steps}

\paragraph{Temporal Standard Deviation Analysis}
\textbf{Purpose:} To distinguish between static objects (furniture, etc.) and dynamic humans

Human presence generates temporal variations in reflection intensity through breathing (\SIrange{0.2}{0.5}{\hertz}) and micro-movements. While static objects maintain constant reflection intensity, regions with humans exhibit temporal variations in reflection intensity due to physiological activities and subtle movements. This processing calculates the temporal variation (standard deviation) of reflection intensity at each position to identify regions indicating human presence.

\paragraph{Adaptive Thresholding and Noise Reduction}
\textbf{Purpose:} To ensure robustness against sensor noise and environmental variations

Detection performance becomes unstable with single fixed thresholds due to environmental condition changes and noise effects. This method uses three-stage adaptive thresholds ($\tau$, $0.8\tau$, $0.6\tau$) and adopts a conservative strategy with maximum value selection to minimize the risk of people count underestimation\cite{adaptiveThresholding}.

\paragraph{Connected Component Analysis and Validation}
\textbf{Purpose:} To accurately count individual people and eliminate false detections

Connected components are extracted through 4-connectivity labeling, and only valid regions corresponding to humans are selected through area and shape constraints\cite{adaptiveThresholding}. By applying validation criteria specialized for low-resolution radar images, noise and abnormally shaped regions are excluded.

\paragraph{Ensemble Prediction}
\textbf{Purpose:} To respond to temporal variation diversity and improve detection accuracy

Processing results from multiple window sizes (10--60~frames) are integrated, and a 30\%-based non-zero detection priority strategy eliminates the effects of incidental noise while capturing reliable human presence signals.

\subsubsection{Detailed Algorithm}

\paragraph{Temporal Standard Deviation Analysis}
Input data consists of 3D data with $12$ rows (range) $\times$ $91$ columns (azimuth) $\times$ $W$ channels (window size), where reflection amplitude values are stored at each pixel position $(x, y)$. For this 3D data, the standard deviation of amplitude values for each pixel in the channel direction (time axis) is calculated, converting to 2D data with 12 rows × 91 columns recording amplitude standard deviations.

To improve spatial coherence, Gaussian smoothing is applied. The Gaussian kernel standard deviation $\sigma_{gaussian} = 0.8$ was adopted after comparing $\{0.5, 0.8, 1.0\}$ in preliminary experiments, considering typical human body size (\SIrange{2}{3}{\pixel} at radar resolution) and achieving optimal signal-to-noise ratio.

\textbf{Computational Complexity:} $O(W \times H \times T)$ (temporal standard deviation) + $O(W \times H)$ (Gaussian smoothing), where $W=12$, $H=91$, $T=$window size.

\paragraph{Adaptive Thresholding and Noise Reduction}

\textit{Multi-threshold Binarization}

For improved detection robustness, ensemble processing is performed using three threshold levels:
\begin{equation}
M_k(x, y) = \begin{cases}
1 & \text{if } \sigma_{smooth}(x, y) > \tau_k \\
0 & \text{otherwise}
\end{cases}
\end{equation}
where $\tau_k \in \{\tau, 0.8\tau, 0.6\tau\}$ and $\tau$ is the primary threshold parameter. This three-stage setting enables gradual detection from strong to weak signals, adopting the result that detects the most human bodies through the maximum value selection strategy described later. Maximum value selection is adopted because people count underestimation (oversight) is practically more serious than overestimation in radar sensing.

\textit{Morphological Operations}

To remove small noise points and non-connected regions in images, morphological operations are applied in the following order:

\textbf{Erosion (1 iteration):} Maintains 1 only when both the pixel and all its 4-neighbors are 1, removing small noise points of \SIrange{1}{2}{\pixel}.

\begin{figure}[htbp]
\centering
\includegraphics[width=0.8\columnwidth]{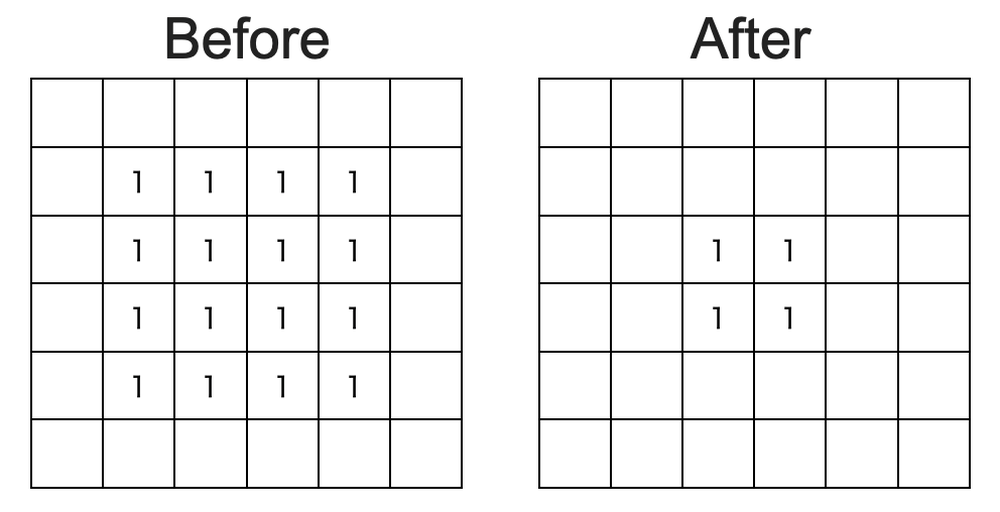}
\caption{Erosion operation example. The original 4×3 rectangular region is reduced to isolated pixels in the center column, demonstrating effective removal of boundary pixels and noise reduction.}
\label{fig:erosion_example}
\end{figure}

As shown in Figure~\ref{fig:erosion_example}, the erosion operation significantly reduces the original connected region. The $4\times3$ rectangular pattern is reduced to only two isolated center pixels, effectively removing boundary pixels and demonstrating the noise reduction capability.

\textbf{Dilation (2 iterations):} Sets a pixel to 1 if either the pixel itself or any of its 4-neighbors is 1. Two iterations restore regions thinned by erosion and connect separated parts.

\begin{figure}[htbp]
\centering
\includegraphics[width=0.8\columnwidth]{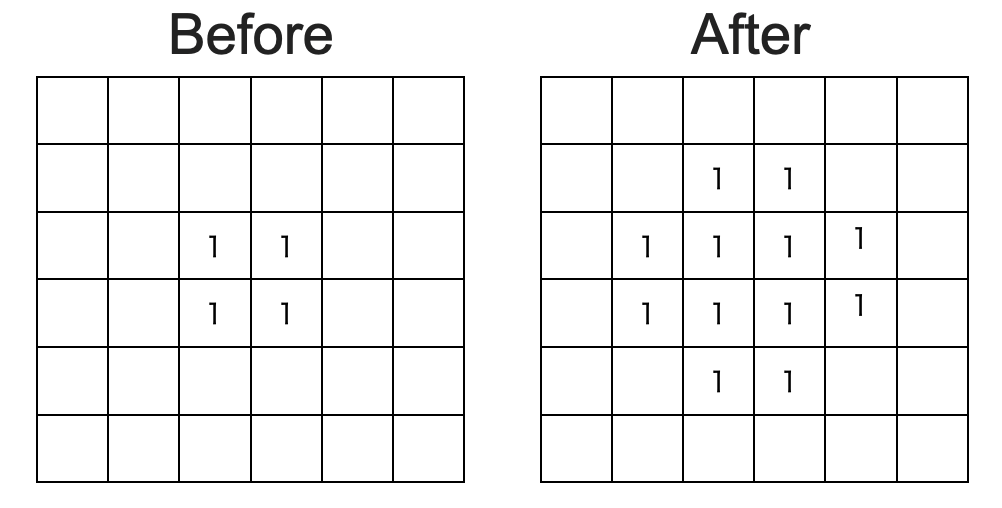}
\caption{Dilation operation example. The isolated center pixels are expanded to form a 3×3 cross pattern, demonstrating region restoration and connectivity enhancement.}
\label{fig:dilation_example}
\end{figure}

Figure~\ref{fig:dilation_example} illustrates the dilation process where the two isolated center pixels expand to form a connected $3\times3$ cross-shaped region. This demonstrates how dilation restores and enlarges regions while maintaining spatial connectivity through 4-neighbor expansion.

The combination of erosion followed by dilation (opening operation) effectively removes small noise while preserving and restoring the main structural features of human body regions in radar images.

\paragraph{Connected Component Analysis and Validation}

\textit{Connected Component Extraction}

Connected components $C_i$ are identified using 4-connectivity labeling. While general computer vision uses 8-connectivity (including diagonal 8-neighbors), radio sensing images have low spatial resolution (12×91), so 8-connectivity risks misrecognizing two or more adjacent human body clusters as one connected component. Preliminary experiments confirmed that 4-connectivity is optimal for individual human body detection.

\textit{Area Constraints}

Area validation is performed for each connected component $C_i$:
\begin{equation}
A_{\min} \leq \operatorname{Area}(C_i) \leq A_{\max}
\end{equation}

$A_{\min} = \SI{2}{\pixel}$: Criterion for excluding 1-pixel isolated point noise. $A_{\max} = \SI{50}{\pixel}$: Set as the maximum area occupied by a single human body, considering radar field of view ($4.9 \times 6.9~\mathrm{m}$) and spatial resolution.

Sensitivity analysis results showed that \SI{50}{\pixel} provided optimal balance when comparing $A_{max}$ values of $\{30, 50, 70\}$.

\textit{Compactness Constraints}

Shape features of each connected component $C_i$ are evaluated:
\begin{equation}
\operatorname{Compactness}(C_i) = \frac{4\pi \, \operatorname{Area}(C_i)}{\operatorname{Perimeter}(C_i)^2}
\end{equation}

Where $\mathrm{Area}(C_i)$ is the area (number of pixels) and $\mathrm{Perimeter}(C_i)$ is the perimeter. Compactness values closer to 1 indicate circular shapes (human-like), while values closer to 0 indicate jagged or elongated shapes (noise).

The threshold 0.1 was adopted after comparing $\{0.05, 0.1, 0.15\}$ in preliminary experiments, achieving optimal noise removal effect.

\paragraph{Ensemble Prediction Strategy}

\textit{Sliding Window Processing}

To respond to diversity in temporal variation patterns, multi-window processing with multiple overlapping windows is performed:

\begin{itemize}
\item \textbf{Window Size:} 
\begin{multline*}
W \in \{ \SI{10}{\frame}, \SI{15}{\frame}, \SI{20}{\frame}, \\
        \SI{25}{\frame}, \SI{30}{\frame}, \SI{60}{\frame} \}
\end{multline*}

Short-term actions (\SI{10}{\frame} $\approx$ \SI{1.2}{\second}) 
to long-term behavioral patterns (\SI{60}{\frame} $\approx$ \SI{7}{\second}) 
are thus captured at multiple temporal scales.

\item \textbf{Step Size:} $\max(1, W/4)$

This ensures 75\% overlap between adjacent windows, 
achieving temporally dense analysis while balancing 
continuity and computational efficiency.
\end{itemize}

\textit{Threshold Ensemble}

Maximum count is selected from three threshold levels for each window:
\begin{equation}
 N_{\mathrm{window}} = \max_k \operatorname{ValidComponents}(M'_k)
\end{equation}

\begin{itemize}
\item $N_{window}$: Final people count for the current window
\item $\mathrm{ValidComponents}(M'_k)$: Number of valid connected components extracted from the $k$-th processed binary map that satisfy area and compactness constraints
\item $\max_k$: Maximum value among results from three threshold levels ($k=1,2,3$)
\end{itemize}

\textit{Temporal Integration}

Robust integration strategy for all window prediction results:

\begin{enumerate}
\item \textbf{Non-zero Detection Priority:} When non-zero detection exceeds 30\% of the total, the mode of non-zero results is adopted. The 30\% threshold is an empirical criterion considering the balance between eliminating incidental noise detection and capturing reliable human presence signals.

\item \textbf{Overall Mode Selection:} When the above condition is not met, the mode from all window results is selected.

\item \textbf{Conservative Tie-breaking:} For multiple values with the same frequency, the higher people count is selected. This is a conservative design because people count underestimation is practically more serious than overestimation.
\end{enumerate}

\textbf{Error Handling:} When no connected components are detected, people count is determined as 0, and boundary pixels are processed with 0-padding.

\subsubsection{Hyperparameter Optimization}

\paragraph{Evaluation Metric Design}
Composite score emphasizing minority classes (1-3 people):
\begin{equation}
\begin{split}
S_{\mathrm{composite}} =\ &0.25 \times \operatorname{Accuracy} + 0.2 \times \mathrm{F1}_{\mathrm{macro}} \\
&+ 0.3 \times \mathrm{F1}_{\mathrm{minority}} + 0.15 \times \mathrm{Recall}_{\mathrm{minority}} \\
&+ 0.1 \times R_{\mathrm{nonzero}}
\end{split}
\end{equation}

\paragraph{Evaluation Metric Definitions}
\begin{itemize}
\item \textbf{Accuracy:} Classification accuracy for all classes (0-3 people)
\item \textbf{F1\_macro:} Macro-average F1 score for all classes (0-3 people)
\item \textbf{F1\_minority:} Macro-average F1 score for classes 1-3
\item \textbf{Recall\_minority:} Macro-average recall for classes 1-3
\item \textbf{R\_nonzero:} Binary classification accuracy (0 people vs. 1+ people)
\end{itemize}

\paragraph{Weighting Priority Rationale}
This weighting reflects practical requirements through the authors' empirical design judgment. Weight order rationale:

\begin{itemize}
\item \textbf{F1\_minority (30\%): Most Important} - Accurate people count identification (1-3 people distinction) is the system's core value, prioritized as an advanced function beyond simple presence detection
\item \textbf{Accuracy (25\%): Second Important} - Basic indicator of overall accuracy essential for user trust
\item \textbf{F1\_macro (20\%): Third Important} - Balance indicator for fair evaluation across classes
\item \textbf{Recall\_minority (15\%): Fourth Important} - Complementary element included in F1\_minority
\item \textbf{R\_nonzero (10\%): Least Important} - Basic function necessary but already evaluated by other indicators and relatively easy to achieve
\end{itemize}

\paragraph{Grid Search Specifications}
Comprehensive parameter exploration:
\begin{itemize}
\item \textbf{Window Size W:} $\{10, 15, 20, 25, 30, 60\}$ frames
\item \textbf{Primary Threshold $\tau$:} $[0.005, 0.08]$ continuous range (50-point sampling)
\end{itemize}

Total exploration points: $6 \times 50 = 300$ combinations

\paragraph{Optimal Parameters}
Parameters selected to maximize $S_{composite}$ through grid search:
\begin{itemize}
\item \textbf{Optimal Window Size:} $W = \SI{20}{\frame}$ ($\approx$ \SI{2.3}{\second})
\item \textbf{Optimal Primary Threshold:} $\tau = 0.025$
\end{itemize}

This combination achieves optimal balance between typical human motion cycles and fine motion detection sensitivity.

\subsection{K-Nearest Neighbors (KNN) Model}

We implemented KNN classifiers using established non-parametric classification principles. The key design focus was on comprehensive statistical feature extraction from radar temporal patterns.

\textbf{Feature Engineering:} For each 60-frame sequence (representing approximately \SI{7}{\second} of radar data), we extracted three fundamental statistical measures: mean reflection intensity, standard deviation (capturing temporal variability), and Gini coefficient (measuring spatial concentration of reflections). To create robust aggregate features, we computed six summary statistics for each measure: median, maximum, minimum, 75th percentile, 25th percentile, and standard deviation, yielding an 18-dimensional feature vector per sequence.

\textbf{Model Configuration:} In this study, we performed a grid search over the combination of three distance metrics—Euclidean, Manhattan, and Mahalanobis—and neighbor counts $k \in \{3, 5\}$, selecting the optimal parameters based on validation performance. For the Mahalanobis distance, a regularized covariance matrix estimated from the training data was used to address potential multicollinearity.

\subsection{Random Forest (RF) Model}

Random Forest was selected as a robust ensemble method that naturally handles feature interactions and provides built-in overfitting resistance through bootstrap aggregation.

\textbf{Feature Engineering:} The model utilized identical 18-dimensional statistical feature vectors as the KNN approach, ensuring consistent feature representation across methods.

\textbf{Model Configuration:} \textbf{Hyperparameter Selection:} Similar to the KNN model, a grid search was conducted to systematically explore parameter combinations. For the tree configurations, the number of estimators $\in \{50, 100\}$ and the maximum depth $\in \{20, \text{None}\}$ were evaluated. These parameters were selected to balance model complexity with computational efficiency, while preventing overfitting on the limited radar dataset.

\subsection{Support Vector Machine (SVM) Model}

SVM implementation leveraged both linear and non-linear decision boundaries to handle the potentially complex separability of radar-derived features.

\textbf{Feature Engineering:} Models employed the same 18-dimensional statistical feature space, maintaining consistency with other classical ML approaches.

\textbf{Model Configuration:} We evaluated linear and radial basis function (RBF) kernels using a grid search to capture both linear and non-linear feature relationships, consistent with the procedure applied to the other classical machine learning methods. The regularization parameter $C = 1.0$ was selected to provide moderate complexity control without extensive hyperparameter tuning, following standard practice for initial SVM deployment.

\subsection{CNN-LSTM Model}

To address the people-counting task, we designed a hybrid neural architecture that integrates a Convolutional Neural Network (CNN) for capturing spatial information with a Bidirectional Long Short-Term Memory (Bi-LSTM) network for modeling temporal dynamics. The model operates on sequences of 60 range–azimuth frames (each of size 12 × 91), where pixel intensities correspond to radar reflection amplitudes.

\begin{figure}[htp]
\centering
\includegraphics[width=1.0\linewidth]{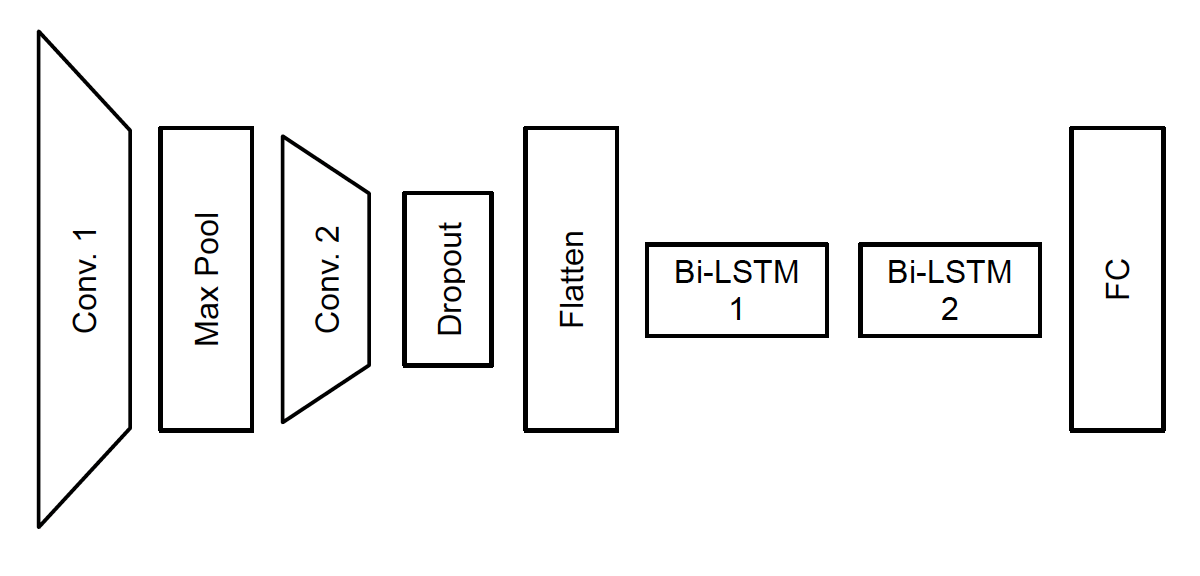}
\caption{Depiction of the CNN–LSTM framework. The convolutional part handles frame-level spatial representation, and the recurrent part captures sequence-level temporal evolution.}
\label{fig:cnn_lstm_arch}
\end{figure}

An overview of the proposed network is presented in Fig.~\ref{fig:cnn_lstm_arch}. For temporal sequence modeling, we utilize a Bi-LSTM structure. This bidirectional design allows simultaneous access to both preceding and subsequent temporal context, which has been demonstrated to yield superior performance compared with unidirectional LSTM variants in radar-based human activity analysis~\cite{biLSTM,continuousClassification}. In radar sensing scenarios, where subtle variations across consecutive frames are critical, bidirectional recurrence enhances the ability to capture complex motion patterns beyond what purely forward-looking models can achieve.

Within the CNN stage, two convolutional layers with ReLU nonlinearities are employed, followed by 2×2 max pooling and dropout (drop rate = 0.3). These operations extract salient spatial features while compressing dimensionality. The resulting feature maps are reshaped into a temporal sequence and passed to a two-layer Bi-LSTM, each direction comprising 128 hidden units. The hidden representation from the final step is subsequently fed into a fully connected (FC) layer to generate a regression output, yielding a continuous-valued estimate of the number of people in the range [0, 3].

The detailed specification of the network is provided in Table~\ref{tab:cnn_lstm_layers}.

\begin{table}[htp]
  \centering
  \small
  \renewcommand{\arraystretch}{1.3}
  \caption{Layer configuration of the baseline CNN-LSTM model.}
  \label{tab:cnn_lstm_layers}
  \begin{tabular}{p{1.3cm}|p{2.6cm}|p{2.6cm}}
    \toprule
    \textbf{Layer}
      & \textbf{Description}
      & \textbf{Output Shape} \\
    \cmidrule(r){1-2}\cmidrule(l){3-3}
    Input
      & 60-frame sequence \newline (13\,$\times$\,92)
      & B\,$\times$\,60\,$\times$\,1\,$\times$\,12\,$\times$\,91 \\
    Conv1
      & Conv2D (1\,$\to$\,16, 3\,$\times$\,3) \newline + ReLU
      & B\,$\times$\,60\,$\times$\,16\,$\times$\,12\,$\times$\,91 \\
    MaxPool
      & MaxPool2D (2\,$\times$\,2)
      & B\,$\times$\,60\,$\times$\,16\,$\times$\,6\,$\times$\,45 \\
    Conv2
      & Conv2D (16\,$\to$\,32, 3\,$\times$\,3) \newline + ReLU
      & B\,$\times$\,60\,$\times$\,32\,$\times$\,6\,$\times$\,45 \\
    Dropout
      & Dropout (p = 0.3)
      & same \\
    Flatten
      & Flatten for LSTM input
      & B\,$\times$\,60\,$\times$\,8640 \\
    Bi-LSTM1
      & Bidirectional LSTM, \newline 128 units (per direction)
      & B\,$\times$\,60\,$\times$\,256 \\
    Bi-LSTM2
      & Bidirectional LSTM, \newline 128 units (per direction)
      & B\,$\times$\,60\,$\times$\,256 \\
    FC
      & Fully connected \newline (256\,$\to$\,1)
      & B\,$\times$\,1 \\
    \bottomrule
  \end{tabular}
\end{table}

Model training was carried out in a regression setting using the Adam optimizer with a learning rate of $10^{-3}$. An early stopping criterion with patience of 10 epochs was employed to prevent overfitting and stabilize convergence.

It should be noted that this network is also employed in a separate manuscript under preparation, where it functions as a baseline model for evaluating spatial generalization strategies. In the current work, by contrast, it is considered as one of several benchmark approaches for comparative analysis.

\section{Experimental Results}

\subsection{Comparison of Classification Accuracy in Environment A}

For evaluation, the Rule-Based Connected Component Model was directly assessed using accuracy, while K-Nearest Neighbors (KNN) Model, Random Forest (RF) Model, Support Vector Machine (SVM) Model, and the CNN-LSTM Deep Learning Model were primarily evaluated using Mean Absolute Error (MAE) and Root Mean Square Error (RMSE). 

MAE summarizes the average size of prediction errors in the original units and treats over- and underestimates equally, indicating how far predictions are from the truth on average. RMSE instead squares errors before averaging and then takes a square root, which magnifies large misses. Thus, RMSE is more sensitive to outliers and better reflects the cost of rare, big errors. In order to provide a fair comparison with the Rule-Based method, the outputs of all learning-based models were rounded to the nearest integer at the first decimal place, and the agreement with the ground-truth labels was computed as classification accuracy.

The classification performance of all five models in Environment A is summarized in Table~\ref{tab:performance_metrics}. 

\begin{table}[htp]
\centering
\caption{Performance Metrics for Environment A}
\label{tab:performance_metrics}
\begin{tabular}{lccc}
\hline
Model & RMSE & MAE & Accuracy \\
\hline
Rule-based Connected Component & -- & -- & 0.4885 \\
Random Forest & 0.3744 & 0.2311 & 0.8260 \\
KNN & 0.3765 & 0.2108 & 0.8385 \\
SVM & 0.4203 & 0.2834 & 0.7771 \\
CNN-LSTM & 0.1544 & 0.0401 & 0.9833 \\
\hline
\end{tabular}
\end{table}

The rule-based approach achieved an accuracy of 48.85\%, indicating limited capability for accurate classification. Traditional machine learning models (KNN, SVM, RF) demonstrated moderate performance with accuracies ranging from 77.71\% to 83.85\%. The deep learning model significantly outperformed all others, achieving 98.33\% accuracy with the lowest RMSE (0.1544) and MAE (0.0401).

Note that the KNN, SVM, and a deep learning model were implemented in this study as regression models, generating continuous-valued outputs rather than discrete class labels. From these continuous predictions, both the MAE and RMSE metrics were computed to capture error characteristics. In order to enable a direct and fair comparison with the rule-based approach, however, the continuous outputs of these regression models were subsequently rounded to the nearest integer value, and the resulting discrete predictions were then evaluated for accuracy by checking their agreement with the ground-truth class labels.

\subsection{Generalization Ability Across Environments}

\subsubsection{Overall Generalization Performance}
To evaluate spatial generalization performance, we applied the five models trained on Environment A to the test data from Environment B. The results are summarized in Table~\ref{tab:generalization_performance}. 

\begin{table}[htp]
\centering
\caption{Generalization Performance on Environment B}
\label{tab:generalization_performance}
\begin{tabular}{lccc}
\hline
Model & RMSE & MAE & Accuracy \\
\hline
Rule-based Connected Component & -- & -- & 0.4919 \\
KNN & 0.7408 & 0.4288 & 0.6312 \\
Random Forest & 0.7018 & 0.4113 & 0.6294 \\
SVM (RBF) & 0.6851 & 0.3919 & 0.6469 \\
CNN-LSTM & 0.6084 & 0.3921 & 0.6200 \\
\hline
\end{tabular}
\end{table}

The rule-based model maintained an accuracy of approximately 49.19\%, similar to its performance in Environment A, indicating consistent and stable performance across different environments. In contrast, the accuracy of the machine learning models dropped significantly from around 80\% to approximately 62--65\%. The deep learning model also experienced a notable decrease, from 98.33\% in Environment A to 62.00\% in Environment B---representing the largest drop among all models.

A detailed inspection of the results revealed that for KNN, RF, SVM, and DL models, the accuracy in estimating the exact number of people (classes 1 to 3) degraded markedly in Environment B. 

\subsubsection{Detailed Analysis of Confusion Matrices}

To understand the degradation patterns in detail, we compare the confusion matrices between Environment A (top) and Environment B (bottom) for each model in Tables~\ref{tab:confusion_matrix}--\ref{tab:confusion_cnn_lstm}.

From the comparison of confusion matrices, it was observed that in Environment~B,
all models exhibited high misclassification rates in multi-person scenarios (classes~1--3),
with particularly pronounced confusion between adjacent classes.
This trend highlights the difficulty of spatial generalization under varying furniture
layouts and spatial configurations.

\begin{table}[htp]
\centering
\caption{Rule-based CC Confusion Matrix Comparison}
\label{tab:confusion_matrix}
\begin{tabular}{|c|c|c|c|c|}
\hline
\multicolumn{5}{|c|}{\textbf{Environment A}} \\
\hline
\textbf{True $\backslash$ Pred} & \textbf{0} & \textbf{1} & \textbf{2} & \textbf{3} \\
\hline
\textbf{0} & 198 & 41 & 1 & 0 \\
\hline
\textbf{1} & 1 & 141 & 71 & 27 \\
\hline
\textbf{2} & 0 & 121 & 78 & 41 \\
\hline
\textbf{3} & 0 & 124 & 64 & 52 \\
\hline
\end{tabular}
\vspace{1ex}

\centering
\begin{tabular}{|c|c|c|c|c|}
\hline
\multicolumn{5}{|c|}{\textbf{Environment B}} \\
\hline
\textbf{True $\backslash$ Pred} & \textbf{0} & \textbf{1} & \textbf{2} & \textbf{3} \\
\hline
\textbf{0} & 374 & 26 & 0 & 0 \\
\hline
\textbf{1} & 42 & 266 & 75 & 17 \\
\hline
\textbf{2} & 0 & 283 & 99 & 18 \\
\hline
\textbf{3} & 2 & 255 & 95 & 48 \\
\hline
\end{tabular}
\end{table}

\begin{table}[htp]
\centering
\caption{Random Forest Confusion Matrix Comparison}
\label{tab:confusion_rf}
\begin{tabular}{|c|c|c|c|c|}
\hline
\multicolumn{5}{|c|}{\textbf{Environment A}} \\
\hline
\textbf{True $\backslash$ Pred} & \textbf{0} & \textbf{1} & \textbf{2} & \textbf{3} \\
\hline
\textbf{0} & 239 & 1 & 0 & 0 \\
\hline
\textbf{1} & 0 & 201 & 38 & 1 \\
\hline
\textbf{2} & 0 & 20 & 180 & 40 \\
\hline
\textbf{3} & 0 & 1 & 66 & 173 \\
\hline
\end{tabular}
\vspace{1ex}

\centering
\begin{tabular}{|c|c|c|c|c|}
\hline
\multicolumn{5}{|c|}{\textbf{Environment B}} \\
\hline
\textbf{True $\backslash$ Pred} & \textbf{0} & \textbf{1} & \textbf{2} & \textbf{3} \\
\hline
\textbf{0} & 400 & 0 & 0 & 0 \\
\hline
\textbf{1} & 21 & 226 & 133 & 20 \\
\hline
\textbf{2} & 0 & 138 & 188 & 74 \\
\hline
\textbf{3} & 0 & 45 & 162 & 193 \\
\hline
\end{tabular}
\end{table}

\begin{table}[htp]
\centering
\caption{KNN Confusion Matrix Comparison}
\label{tab:confusion_knn}
\begin{tabular}{|c|c|c|c|c|}
\hline
\multicolumn{5}{|c|}{\textbf{Environment A}} \\
\hline
\textbf{True $\backslash$ Pred} & \textbf{0} & \textbf{1} & \textbf{2} & \textbf{3} \\
\hline
\textbf{0} & 238 & 1 & 1 & 0 \\
\hline
\textbf{1} & 0 & 199 & 37 & 4 \\
\hline
\textbf{2} & 0 & 11 & 181 & 48 \\
\hline
\textbf{3} & 0 & 1 & 52 & 187 \\
\hline
\end{tabular}
\vspace{1ex}

\centering
\begin{tabular}{|c|c|c|c|c|}
\hline
\multicolumn{5}{|c|}{\textbf{Environment B}} \\
\hline
\textbf{True $\backslash$ Pred} & \textbf{0} & \textbf{1} & \textbf{2} & \textbf{3} \\
\hline
\textbf{0} & 400 & 0 & 0 & 0 \\
\hline
\textbf{1} & 18 & 231 & 110 & 41 \\
\hline
\textbf{2} & 0 & 127 & 180 & 93 \\
\hline
\textbf{3} & 0 & 55 & 146 & 199 \\
\hline
\end{tabular}
\end{table}

\begin{table}[htp]
\centering
\caption{SVM Confusion Matrix Comparison}
\label{tab:confusion_svm}
\begin{tabular}{|c|c|c|c|c|}
\hline
\multicolumn{5}{|c|}{\textbf{Environment A}} \\
\hline
\textbf{True $\backslash$ Pred} & \textbf{0} & \textbf{1} & \textbf{2} & \textbf{3} \\
\hline
\textbf{0} & 238 & 2 & 0 & 0 \\
\hline
\textbf{1} & 0 & 185 & 53 & 2 \\
\hline
\textbf{2} & 0 & 26 & 158 & 56 \\
\hline
\textbf{3} & 0 & 3 & 72 & 165 \\
\hline
\end{tabular}

\vspace{1ex}

\centering
\begin{tabular}{|c|c|c|c|c|}
\hline
\multicolumn{5}{|c|}{\textbf{Environment B}} \\
\hline
\textbf{True $\backslash$ Pred} & \textbf{0} & \textbf{1} & \textbf{2} & \textbf{3} \\
\hline
\textbf{0} & 400 & 0 & 0 & 0 \\
\hline
\textbf{1} & 18 & 228 & 129 & 25 \\
\hline
\textbf{2} & 0 & 146 & 183 & 71 \\
\hline
\textbf{3} & 0 & 37 & 139 & 224 \\
\hline
\end{tabular}
\end{table}

\begin{table}[htp]
\centering
\caption{CNN-LSTM Confusion Matrix Comparison}
\label{tab:confusion_cnn_lstm}
\begin{tabular}{|c|c|c|c|c|}
\hline
\multicolumn{5}{|c|}{\textbf{Environment A}} \\
\hline
\textbf{True $\backslash$ Pred} & \textbf{0} & \textbf{1} & \textbf{2} & \textbf{3} \\
\hline
\textbf{0} & 239 & 0 & 1 & 0 \\
\hline
\textbf{1} & 0 & 233 & 6 & 1 \\
\hline
\textbf{2} & 0 & 2 & 237 & 1 \\
\hline
\textbf{3} & 0 & 1 & 4 & 235 \\
\hline
\end{tabular}

\vspace{1ex}

\centering
\begin{tabular}{|c|c|c|c|c|}
\hline
\multicolumn{5}{|c|}{\textbf{Environment B}} \\
\hline
\textbf{True $\backslash$ Pred} & \textbf{0} & \textbf{1} & \textbf{2} & \textbf{3} \\
\hline
\textbf{0} & 400 & 0 & 0 & 0 \\
\hline
\textbf{1} & 24 & 213 & 163 & 0 \\
\hline
\textbf{2} & 0 & 133 & 177 & 90 \\
\hline
\textbf{3} & 0 & 6 & 192 & 202 \\
\hline
\end{tabular}
\end{table}


\subsubsection{Binary Classification Accuracy (No Person vs.\ Person Present)}

To further evaluate the models’ ability to detect the mere presence of a person, we collapse the four original classes into two: “No Person” (class 0) and “Person Present” (classes 1–3).  We then compute the binary accuracy as:

\begin{align}
\mathrm{Accuracy}_{\mathrm{binary}}
&= \frac{\mathrm{TN} + \mathrm{TP}}
       {\mathrm{TN} + \mathrm{FP} + \mathrm{FN} + \mathrm{TP}} \nonumber\\
&= \frac{C_{0,0} + \displaystyle\sum_{i=1}^{3}\sum_{j=1}^{3} C_{i,j}}
       {\displaystyle\sum_{i=0}^{3}\sum_{j=0}^{3} C_{i,j}}.
\end{align}

where
\begin{itemize}
  \item $C_{i,j}$ is the $(i,j)$‐entry of the $4\times4$ confusion matrix,
  \item $\mathrm{TN}=C_{0,0}$,
  \item $\mathrm{FP}=\sum_{j=1}^{3}C_{0,j}$,
  \item $\mathrm{FN}=\sum_{i=1}^{3}C_{i,0}$,
  \item $\mathrm{TP}=\sum_{i=1}^{3}\sum_{j=1}^{3}C_{i,j}$.
\end{itemize}

\vspace{1em} 

The recalculated binary classification accuracy is summarized in Table~\ref{tab:binary_accuracy}. 
Notably, when the output granularity was reduced from four-class occupancy estimation (0--3 persons) 
to binary classification of ``No Person'' versus ``Person Present,'' all five models demonstrated strong 
spatial generalization ability. In particular, even under the altered spatial layout of Environment~B, 
each model achieved an accuracy exceeding 95\%.

\begin{table}[htp]
  \centering
  \caption{Binary Classification (No Person vs.\ Person Present) Accuracy}
  \label{tab:binary_accuracy}
  \begin{tabular}{lcc}
    \hline
    \textbf{Model}                    & \textbf{Env A} & \textbf{Env B} \\
    \hline
    Rule-based Connected Component    & 95.52\%        & 95.63\%        \\
    Random Forest                     & 99.90\%        & 98.69\%        \\
    KNN              & 99.79\%        & 98.88\%        \\
    SVM                         & 99.79\%        & 98.88\%        \\
    CNN–LSTM                          & 99.90\%        & 98.50\%        \\
    \hline
  \end{tabular}
\end{table}

\section{Discussion}
\subsection{Accuracy in a Fixed Environment}
In Environment~A, the CNN--LSTM model achieved the highest accuracy (98.33\%), 
followed by KNN (83.85\%), Random Forest (82.60\%), SVM (77.71\%), 
and the rule-based method (48.85\%). 
This result suggests a trend in which higher model complexity leads to improved classification performance 
within the same environment. 
In particular, the exceptionally high accuracy of the CNN--LSTM model can be attributed to the complementary 
characteristics of CNNs and LSTMs: CNNs extract spatial features from amplitude values, while LSTMs capture 
temporal variations, and their integration enables highly effective discrimination.  

In contrast, despite being designed on the basis of a radar-based people counting method that combines 
temporal variation analysis (peak detection) widely used in signal processing with connected component 
analysis established in computer vision, the rule-based approach yielded the lowest accuracy. 
This indicates that when the objective is relatively fine-grained outputs such as estimating the number 
of people (0--3), rule-based methods with linear characteristics face inherent limitations.

\subsection{Performance under Spatial Generalization}
When transitioning to Environment~B, which had a different background layout, all learning-based models (KNN, RF, SVM, CNN--LSTM) experienced substantial performance drops of 13.02--36.33 percentage points. In contrast, the rule-based method slightly improved from 48.85\% to 49.19\%. This behavior suggests that while high-capacity models are effective in fixed environments, they tend to overfit to background patterns in the training environment and lose robustness when confronted with unseen spatial configurations. 

In real-world sensing environments, the \emph{i.i.d.} (independent and identically distributed) assumption—that each sample is statistically independent and drawn from the same underlying distribution—rarely holds; input distributions can easily change due to factors such as furniture relocation, weather variations, or sensor degradation~\cite{CrossEnvElectronics2025,Hernangomez2022,Khodabakhshandeh2021}. 
Therefore, the deployment of learning-based models in environments with large input distribution shifts should be conducted with caution. 
It is also noteworthy that in our experiments, the CNN--LSTM model, which achieved the highest accuracy in the same environment, exhibited the largest performance drop when applied to Environment~B with a different background layout. 
This finding suggests the existence of a trade-off between achieving high discriminative performance in a fixed environment and maintaining spatial generalization ability. 
In general, as the representational capacity of a model increases, its ability to fit the training data also improves; however, this comes at the cost of reduced adaptability to unseen environments, making such a trade-off essentially inevitable.

\subsection{Output Granularity vs.\ Spatial Generalization}
By simplifying the classification task from a fine-grained four-class problem (0--3 persons) to a coarse-grained 
binary task (person present vs.\ no person), all models achieved accuracies exceeding 95\% even in Environment~B, 
which had a different background layout. This result suggests the existence of a trade-off between output granularity 
and spatial generalization ability (see also \cite{GroupedPeopleCounting2023} for a representative multi-class people counting task). 

In other words, maintaining high fine-grained classification performance while ensuring spatial generalization 
remains a significant challenge. 

\subsection{Guidelines for Practical Deployment}

Considering practical deployment in dynamic physical environments, the following design guidelines are derived from our results:
\begin{itemize}
\item High-capacity models such as deep learning should be deployed only in environments with relatively stable layouts, or where continuous model updates via transfer learning or test-time adaptation are feasible, in order to mitigate performance degradation under domain shift~\cite{Hernangomez2022,Khodabakhshandeh2021,CrossEnvElectronics2025}.
\item In scenarios where layout changes occur frequently and adaptation through transfer learning is difficult, a simple binary output design (e.g., presence detection) should be adopted, as it is inherently robust to environmental variations.
\item Preprocessing techniques that extract domain-invariant features should be introduced to further enhance robustness against environmental changes. In this study, sigmoid-based amplitude weighting was employed; however, in other domains or sensing modalities, the optimal preprocessing method should be selected according to application-specific characteristics~\cite{adaptiveThresholding,hybridNeural,targetRecognition}.
\end{itemize}

These guidelines provide concrete and actionable insights for the implementation of people-counting and presence-detection systems in real-world environments within the Radio Wave Sensing domain. In particular, they offer direct guidance on diverse design choices such as model selection, the determination of output granularity, and the adoption of adaptation strategies to maintain performance under changing environmental conditions. As a result, these guidelines contribute to ensuring stable reliability across different spaces and to the construction of a robust sensing framework suitable for practical deployment in real-world applications.

\section{Conclusion}

This study investigated the trade-offs among accuracy, spatial generalization, and output granularity 
for human estimation using radio wave sensing. Experiments conducted across two different environments 
demonstrated that high-capacity models such as CNN--LSTM achieved extremely high accuracy within the 
same environment, but their robustness degraded substantially under domain shift, such as changes in 
spatial layout. This finding indicates an inherent trade-off between the ability to achieve fine-grained 
outputs (e.g., multi-class headcount estimation) and the ability to generalize across environments.  

In contrast, the rule-based approach yielded relatively low accuracy for fine-grained headcount estimation, 
yet maintained stable performance across different environments. Furthermore, when the output was simplified 
to binary classification (i.e., presence vs.\ absence), all models, including learning-based approaches, 
achieved accuracies exceeding 95\%. These results highlight that output granularity is a critical factor in 
determining robustness to spatial variation, and that binary designs inherently provide resilience in 
dynamic environments.  

The insights obtained from this study provide concrete guidelines for constructing flexible and robust 
sensing systems that balance accuracy, robustness, and computational efficiency.  

\section*{Acknowledgment}
The authors would like to thank Prof. Tei Sigaku, Emeritus Professor at the University of Aizu, and Dr. Aisaku Nakamura for their valuable guidance and discussions. The authors also thank SoftBank Corp. for providing the research environment and technical support during the experiments.



\begin{thebibliography}{00}

\bibitem{Kong2025}
H. Kong, S. Shen, S. Li, \emph{et al.}, ``A survey of mmWave radar-based sensing in autonomous vehicles, smart homes and industry,'' \textit{IEEE Commun. Surv. Tutor.}, 2025.

\bibitem{millimeterReview}
F. Lemic, A. Behboodi, V. Handziski, and A. Wolisz, ``A review of millimeter wave device-based localization and device-free sensing technologies and applications,'' \textit{IEEE Commun. Surv. Tutor.}, vol. 24, no. 3, pp. 1679--1718, 2022.

\bibitem{frequencyTracking}
F. Fioranelli, \emph{et al.}, ``A novel frequency-tracking algorithm for noncontact vital sign monitoring,'' \textit{IEEE Sens. J.}, vol. 23, no. 18, pp. 20782--20793, 2023.

\bibitem{SensorsVehOcc2024}
W. Li, C. Wang, Y. Zhang, \emph{et al.}, ``Vehicle occupant detection based on mm-wave radar,'' \textit{Sensors}, vol. 24, no. 11, 3334, 2024.

\bibitem{SmartBldgMultiRadar2024}
V. Barral, T. Domínguez-Bolaño, C. J. Escudero, and J. A. García-Naya, ``An IoT system for smart building combining multiple mmWave FMCW radars applied to people counting,'' arXiv:2401.17949, 2024.

\bibitem{radarays}
A. Mock, M. Magnusson, and J. Hertzberg, ``RadaRays: Real-time simulation of rotating FMCW radar for mobile robotics via hardware-accelerated ray tracing,'' \textit{IEEE Robot. Autom. Lett.}, 2025.

\bibitem{adaptiveThresholding}
A. Seifert \emph{et al.}, ``Radar-based human activity recognition with adaptive thresholding towards resource constrained platforms,'' \textit{Sci. Rep.}, vol. 13, 4331, 2023.

\bibitem{targetRecognition}
X. Li \emph{et al.}, ``Radar target characterization and deep learning in radar automatic target recognition: A review,'' \textit{Remote Sens.}, vol. 15, no. 15, 3742, 2023.

\bibitem{humanMotion}
S. Z. Gurbuz and M. G. Amin, ``Radar-based human-motion recognition with deep learning: Promising applications for indoor monitoring,'' \textit{IEEE Signal Process. Mag.}, vol. 36, no. 4, pp. 16--28, 2019.

\bibitem{hybridCNN}
J. Zhu, H. Chen, and W. Ye, ``A hybrid CNN–LSTM network for the classification of human activities based on micro-Doppler radar,'' \textit{IEEE Access}, vol. 8, pp. 24713--24720, 2020.

\bibitem{deepLearning}
T.-H. Tan, J.-H. Tian, A. K. Sharma, S.-H. Liu, and Y.-F. Huang, ``Human activity recognition based on deep learning and micro-Doppler radar data,'' \textit{Sensors}, vol. 24, no. 8, 2530, 2024.

\bibitem{CrossEnvElectronics2025}
R. El Hail, P. Mehrjouseresht, D. M. M. P. Schreurs, and P. Karsmakers, ``Radar-based human activity recognition: A study on cross-environment robustness,'' \textit{Electronics}, vol. 14, no. 5, 2025.

\bibitem{ourPriorWork}
T. Tanaka, A. Yabuki, M. Funakoshi, and R. Yonemoto, ``Validation of practicality for CSI sensing utilizing machine learning,'' in \textit{Intelligent Human Computer Interaction}. Cham, Switzerland: Springer, 2025, pp. 243--257, doi: 10.1007/978-3-031-92605-1\_16.

\bibitem{microwaveSurveyCorrect}
I. Ullmann, R. G. Guendel, N. C. Kruse, F. Fioranelli, and A. Yarovoy, ``A survey on radar-based continuous human activity recognition,'' \textit{IEEE J. Microw.}, vol. 3, pp. 938--950, 2023.

\bibitem{passiveRadar}
O. Simeone, \emph{et al.}, ``Passive radar sensing for human activity recognition: A survey,'' \textit{IEEE Commun. Surv. Tutor.}, vol. 26, no. 3, pp. 1678--1720, 2024.

\bibitem{Hernangomez2022}
R. Hernangómez, I. Bjelaković, L. Servadei, and S. Stańczak, ``Unsupervised domain adaptation across FMCW radar configurations using margin disparity discrepancy,'' in \textit{Proc. Eur. Signal Process. Conf. (EUSIPCO)}, 2022, pp. 1--5.

\bibitem{Khodabakhshandeh2021}
H. Khodabakhshandeh, T. Visentin, R. Hernangómez, and M. Pütz, ``Domain adaptation across configurations of FMCW radar for deep learning based human activity classification,'' in \textit{Proc. Int. Radar Symp. (IRS)}, 2021, pp. 1--10.

\bibitem{GroupedPeopleCounting2023}
W. Ren, A. Yarovoy, and F. Fioranelli, ``Grouped people counting using mm-wave FMCW MIMO radar,'' \textit{IEEE Internet Things J.}, vol. 10, no. 24, pp. 21962--21973, 2023.

\bibitem{exactLinear}
A. Bartsch \emph{et al.}, ``A more exact linear FMCW radar signal model for simultaneous range-velocity estimation,'' in \textit{Proc. IEEE Radar Conf. (RadarConf18)}, 2018, pp. 532--537.

\bibitem{ieiceRadarTechRep}
T. Tanaka, R. Yonemoto, M. Funakoshi, and A. Yabuki, ``Development of radio wave sensing system combining radar and CNN/LSTM machine learning models: Achieving high-accuracy people counting with small training data,'' \textit{IEICE Tech. Rep.}, vol. 125, no. 1, AP2025-5, pp. 24--29, Apr. 2025.

\bibitem{biLSTM}
H. Li, A. Shrestha, H. Heidari, J. Le Kernec, and F. Fioranelli, ``Bi-LSTM network for multimodal continuous human activity recognition and fall detection,'' \textit{IEEE Sens. J.}, vol. 20, no. 3, pp. 1191--1201, 2020.

\bibitem{continuousClassification}
A. Shrestha \emph{et al.}, ``Continuous human activity classification from FMCW radar with Bi-LSTM networks,'' \textit{IEEE Sens. J.}, vol. 20, no. 22, pp. 13607--13619, 2020.

\bibitem{hybridNeural}
W. Ding \emph{et al.}, ``Radar-based human activity recognition using hybrid neural network model with multidomain fusion,'' \textit{IEEE Trans. Aerosp. Electron. Syst.}, vol. 57, no. 5, pp. 2889--2898, 2021.

\end{thebibliography}
\end{document}